\documentclass[runningheads]{llncs}
\usepackage[T1]{fontenc}
\usepackage{graphicx}
\usepackage{xcolor}
\usepackage{float}
\usepackage{devanagari}

\begin{document}
\title{HindiLLM: Large Language Model for Hindi}
%
%
\author{Sanjay Chouhan\inst{1} \and
Shubha Brata Nath\inst{1} \and
Aparajita Dutta\inst{2}}
\authorrunning{S. Chouhan et al.}
%
\institute{Indian Institute of Information Technology Guwahati, Bongora, Assam 781015, India\\
\email{\{sanjay.chouhanm22,shubha\}@iiitg.ac.in}\\
\and
National Institute of Technology Silchar, Silchar, Assam 788010, India\\
\email{aparajita.dutta@cse.nits.ac.in}}
\maketitle              
\begin{abstract}
The advancements in the Large Language Model (LLM) have helped in solving several problems related to language processing. Most of the researches have focused on the English language only, because of its popularity and abundance on the internet. However, a high-performance language model for Hindi and other Indic languages is lacking in the literature. In this work, we have pre-trained two autoregressive LLM models for the Hindi language, namely HindiLLM-Small and HindiLLM-Medium. We use a two-step process comprising unsupervised pre-training and supervised fine-tuning. First, we create a large and high-quality text corpus for unsupervised pre-training. Next, we train a Byte-Pair Encoding, named HindiLLM tokenizer, using the pre-training text data. We then perform training on the unlabeled data, known as the pre-training step, to get the HindiLLM base models. Furthermore, we perform fine-tuning of the HindiLLM base models for different tasks like sentiment analysis, text classification, natural language inference, and multiple choice question-answer on popular labeled datasets to measure the real-world performance. The evaluation shows that the HindiLLM-based fine-tuned models outperform several models in most of the language related tasks.

\end{abstract}

\section{Introduction}
\label{introduction}

The understanding of a language by a machine is of great interest in the Natural Language Processing (NLP) domain. The words and phrases used in a sentence can have different meanings in different contexts of a sentence. Also, there is the need to identify synonyms of a word, sarcastic phrases, idioms, and errors in the text of a language. The language models in NLP can perform these tasks for the English language as there is a rich collection of datasets in the literature and it has been highly researched over the years. Most of the other languages, especially Indo-Aryan languages or Indic languages, are less researched and have fewer resources as their presence on the internet is limited.

The Large Language Models (LLMs) are state-of-the-art for NLP tasks and have been able to show great progress regarding language comprehension, inference, and other language analysis. The language-related tasks for English are processed by LLMs such as GPT-4 \cite{achiam2023gpt} (a model from the GPT series), LLaMA-2 \cite{touvron2023llama} (a model from the LLaMA series), PaLM-2 \cite{anil2023palm} (a model from the PaLM series) and Mistral-7B \cite{jiang2023mistral}. These models have billions of parameters and are trained on hundreds of GBs of data. Hence, they can perform advanced NLP tasks such as instruction following, code generation, new content generation, and information retrieval. However, the initial versions of these models were smaller and hence less capable. 

Hindi is an Indic language written in the Devanagari script. It is a subject-object-verb (SOV) language and is morphologically rich. According to the $26^{th}$ edition of Ethnologue\footnote[1]{\url{https://www.ethnologue.com/insights/ethnologue200/}} published in 2023, Hindi is the third most spoken language having 609.5 million speakers. It is the official language of India and according to Forbes India\footnote[2]{\url{https://www.forbesindia.com/article/news-by-numbers/hindi-day-2020-indias-mostspoken-languages-are/62577/1}}, 44\% of India speaks Hindi. However, the language models for the Hindi language lag behind due to the challenges mentioned below.

\begin{itemize}
    \item The data collection has issues as there is limited availability of rich Hindi corpora, both for pre-training as well as fine-tuning.
    \item The complexity of Hindi Devanagari text needs to be taken into consideration. The intricacies of the Hindi script including conjunct characters and nuanced linguistic structures introduce complexity in Hindi text processing. 
    \item The NLP model needs to understand the Hindi language in various contexts for tasks like sentiment analysis, inference and text summarization.
\end{itemize}

This work aims at the following contributions to tackle the above mentioned challenges.

\begin{itemize}
    \item Our work focuses on training a tokenizer for the Hindi language using Byte Pair Encoding (BPE) tokenization algorithm\footnote[3]{\url{https://huggingface.co/learn/nlp-course/en/chapter6/5}}.
    \item We train two autoregressive models for the Hindi language of different sizes. 
    \item We perform supervised fine-tuning for multiple downstream tasks of the Hindi language. \end{itemize}

The remainder of the paper is organized as follows. We discuss the existing works in Section \ref{related_work}. Section \ref{dataset} explains the dataset used in this work. We discuss the methodology in Section \ref{training_approach}. The performance evaluation is presented in Section \ref{evaluation}. Finally, Section \ref{conclusion} concludes the work with future directions.

\section{Related Work}
\label{related_work}

There is a scarcity of research on the language models for Hindi language and other Indic languages. Hindi speakers are present all over the world and yet it is less-researched in the field of NLP. One major reason behind this is the limited presence of Indic languages online in written form. Now, we discuss the related works present in the literature.

In the work by Arora et al. \cite{arora2020inltk}, the authors focused on pre-training language models for 13 Indic languages. They pre-trained ULMFiT \cite{howard2018universal} and TransformerXL \cite{dai2019transformer} language models from scratch. The authors in Kakwani et al. \cite{kakwani2020indicnlpsuite} contributed multiple resources such as monolingual corpora, pre-trained word embeddings, IndicGLUE benchmark dataset and pre-trained language models. They used the ALBERT \cite{lan2019albert} model for pre-training language models. ALBERT, being a compact model, is lightweight and requires less training data which is good for less-resource languages. 

Vries et al. \cite{de2020good} recycled the pre-trained GPT-2 \cite{radford2019language} models for Italian and Dutch languages. They primarily retrained the lexical embeddings. First, they created a new BPE vocabulary for the target language. Then, they re-initialized the lexical embeddings of the GPT-2 model for the new vocabulary and re-trained them. They  mentioned that the full model can be finetuned later with a smaller learning rate. This helps the model to better adjust to the new language whereas the re-learned lexical embeddings reduce the risk of information loss. 

Owen et al. \cite{owen2024komodo} discussed incremental pre-training for adapting English based LLMs to non-English languages such as Indonesian. First, they expanded the vocabulary by integrating a new trained tokenizer with the existing one. Then, they did incremental pre-training of Llama-2-7B-Base \cite{touvron2023llama} using Low-Rank Adaptation (LORA) \cite{hu2021lora} technique. This helps the model to learn new language without catastrophically forgetting the English language in a minimal resource requirement setup. Owen et al. \cite{owen2024komodo} and Vries et al. \cite{de2020good} discussed about utilizing English based pre-trained LLMs. Since these are pre-trained on English language, their techniques are not suitable for building the HindiLLM models because Hindi is very different from English and is written in Devanagari instead of Latin.

Niyogi et al. \cite{niyogi2024paramanu} trained multiple auto-regressive models from scratch for $10$ Indian languages. Their models are also based on Transformers decoders. However, they did not provide a detailed description of the data used and the model architecture. Radford et al. \cite{radford2018improving} discussed semi-supervised training approach which is a two-step training process for the English language. The first step is generative pre-training on a large unlabeled corpus which gives a good initialization point for the next step, which is discriminative fine-tuning on each specific task. The GPT-1 is an auto-regressive transformer \cite{vaswani2017attention} based decoder only model. We utilized similar semi-supervised training approach for our HindiLLM models.

Although our work focuses on the Hindi language, we can apply the language model techniques across different languages. This will enable the less-researched languages to benefit by utilizing the techniques mentioned for extensively researched languages.

\section{Dataset}
\label{dataset}

The dataset generation is the first step of any model building process. The richness of the data determines the quality of the model. The approach used in this paper requires two types of data, unlabeled data for unsupervised pre-training and labeled data for supervised fine-tuning. The details of the datasets used in this paper are mentioned in this section.

\subsection{Pre-Training Dataset}
\label{dataset-pre_training}
For the unsupervised pre-training step, we need a large corpus with Hindi text written in Devanagari script. Since pre-training is the most crucial step which helps the model in understanding the language and its nuances, we need the corpus to be clean and have valid long paragraphs. This implies that it should not have unnecessary symbols, words, web-links and characters which we do not see in typical Hindi literature. A paragraph or sentence is valid if it follows the grammar of that language and it makes sense to the native speaker of that language. Long paragraphs are good for introducing long-term dependencies in the model. Generally, we need to scrap the internet for large corpus but this leads to difficulty in finding usable Hindi sentences or paragraphs. There exist projects which focus on getting web-crawled texts. In these projects, these corpora are classified based on language and are preprocessed following the structure of that language. We have used such existing corpora for the pre-training step. 

\begin{table}
    \centering
    \caption{Detailed Description of Pre-training Corpora}
    \label{tab:pre-training_data_description}
    \begin{tabular}{|l|r|r|} \hline  
         \textbf{Data}&  \textbf{Size
\textit{(in GB)}}& \textbf{No. of Words
\textit{(approx. in million)}}\\ \hline  
         Wikipedia&  1.04& 78.82\\ \hline  
         CC-Aligned&  1.3& 133.89\\ \hline  
         OSCAR-2201&  14& 1185.51\\ \hline  
         CC-100&  21& 1714.72\\ \hline 
    \end{tabular}
\end{table}

As shown in Table \ref{tab:pre-training_data_description}, we have 37.34 GB of data which contains approximately 3.11 billion words. Apart from CC-Aligned \cite{el2019ccaligned}, all other datasets contain only Hindi text written in Devanagari script. The CC-Aligned dataset contains Hind-English translation pairs of sentences and phrases. We have concatenated both versions (Hindi and English) of the sentence one after the other. One of the sentences from the translation pair is chosen randomly as the first sentence for each of the translation pairs. The idea here is to add some English capability in the model along with Hindi because we often see words or phrases of English inserted in Hindi texts. The translation pair will also help the model understand the relationship between both languages. Hindi Wikipedia articles are also used from Kaggle\footnote[4]{\url{https://www.kaggle.com/datasets/disisbig/hindi-wikipedia-articles-172k}} and Tensorflow\footnote[5]{\url{https://www.tensorflow.org/datasets/catalog/wikipedia}}. Wikipedia articles contain factual information and the sentences are well formed as well, unlike some internet forums. The OSCAR-2201 \cite{abadji2022towards} or Open Super-large Crawled Aggregated coRpus is a multilingual corpus intended for pre-training language models. It contains 151 different languages but we only use the Hindi texts. The CC-100 \cite{conneau2019unsupervised} \cite{wenzek2019ccnet} is a monolingual corpus containing texts in 100+ languages out of which we only utilize the Hindi texts. It is generated using the open-source CC-Net \cite{wenzek2019ccnet} repository. 

Apart from the default preprocessing done by the dataset creator, we have additionally performed the following preprocessing steps.
\begin{itemize}
    \item \textbf{Content Cleanup} - Removal of content within brackets, hyperlinks, extra spaces and formatting of punctuation marks.
    \item \textbf{Filtering Sentence} - Sentences where less than 30\% of words are unique are filtered out. These are not valid sentences because same few words are repeated multiple times.
    \item \textbf{Numeric and Punctuation Removal} - Removal of lines containing only numeric, punctuation marks or special symbols.
    \item \textbf{Handling Short Lines} - Multiple consecutive lines with less than fours words were removed. These were generally navigation links of a website.
\end{itemize}

\subsection{Fine-Tuning Dataset}
\label{dataset-fine_tuning}
The fine-tuning is done for downstream tasks. The performance on downstream tasks tells about the real-world applicability of the model. In this paper, we have chosen seven downstream tasks to measure different aspects of our models. 

\subsubsection{Sentiment Analysis:} We have two sentiment analysis datasets, namely IITP Movie\footnote[6]{\url{https://www.kaggle.com/datasets/warcoder/iit-patna-movie-reviews-hindi}} and IITP Product\footnote[7]{\url{https://www.kaggle.com/datasets/warcoder/iit-patna-product-reviews}}. These datasets are public and widely used for evaluating Hindi language models. The dataset comprises three classes: positive, neutral, and negative. We have combined both datasets for training but tested separately.

\subsubsection{Text Classification:} For multiclass classification evaluation, we have used BBC News category\footnote[8]{\url{https://github.com/NirantK/hindi2vec/releases/tag/bbc-hindi-v0.1}} classification dataset. It has six categories. These are India, international, news, entertainment, sports, and science. It is also a public dataset which has been used for testing multiple Hindi based language models.

\subsubsection{Natural Language Inference:} We analyzed the natural language inference capability of our models with the BBC NLI \cite{uppal2020two} dataset. 


\subsubsection{Cloze-style Multiple-choice QA (CSQA):} The CSQA dataset is from IndicGLUE \cite{kakwani2020indicnlpsuite} benchmark dataset. This dataset has a masked entity in a given text and we are given four candidate entities out of which one is the correct entity. This dataset is created from Wikipedia articles.

\subsubsection{Wikipedia Section-title Prediction (WSTP):} Similar to CSQA, WSTP is from the IndicGLUE benchmark dataset and created using Wikipedia articles. The dataset has Wikipedia sections and it is required to find out the section title from the given four choices of titles. 

\subsubsection{Discourse Mode Classification (DM):} The DM dataset is also from the IndicGLUE benchmark dataset. It is a discourse analysis dataset with five discourse categories: descriptive, narrative, dialogue, argumentative and informative. In this task, the model has to predict the suitable discourse category for a given sentence.

\subsubsection{Machine Translation Dataset:} We have obtained the translation dataset from Kunchukuttan et al. \cite{kunchukuttan2017iit}. It has pairs of Hindi and English versions of 1.49 million sentences. This dataset is specifically created for the Hindi-English translation task. The machine translation is a generative downstream task.

\section{Methodology}
\label{training_approach}
We follow two steps training process in this work. We apply an unsupervised pre-training to get the base model and a supervised fine-tuning of the base model for the downstream tasks. However, prior to the training, we build the HindiLLM tokenizer using the BPE algorithm. In this section, we have provided a detailed description of the tokenizer and the models along with the training process.

\subsection{Tokenizer}
\label{training_approach-tokenizer}
Since we focus on building the model for the Hindi language and the default GPT-2 \cite{radford2019language} model is primarily for the English language. So, we train a new tokenizer called HindiLLM tokenizer. The idea of training a custom tokenizer is to reduce the fertility score (average number of tokens per word) for the Hindi language. We have trained a Byte-level BPE tokenizer with our pre-training corpora which contains mostly Hindi language written in Devanagari script. Since the Devanagari script is complex, the Byte-level BPE tokenizer is the most suitable option. We have used the whole pre-training dataset to train the BPE tokenizer. We have kept the desired vocabulary size as 50000 while using the trainer to accommodate most of the frequently occurring sub-words. Also, we added 8 special tokens like CLS, SEP and PAD afterwards, so the vocabulary size reached to 50008.

\begin{figure}
    \centering
    \includegraphics[width=\columnwidth]{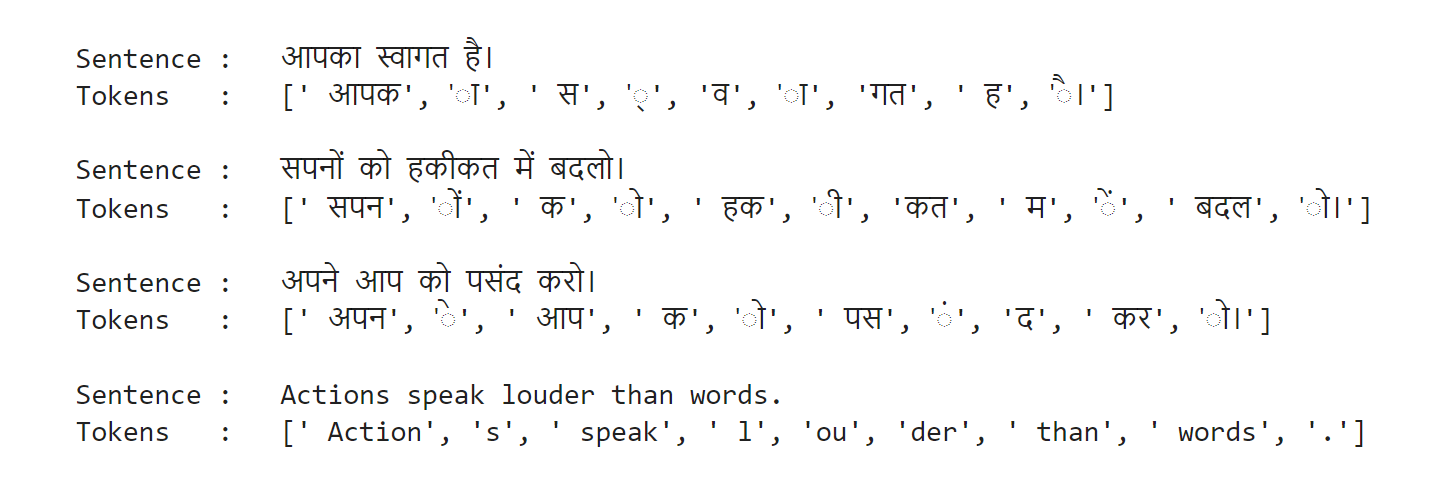}
    \caption{Output of HindiLLM Tokenizer}
    \label{fig:HindiLLM-Tokenizer_output}
\end{figure}

We show the output of the HindiLLM tokenizer in Figure \ref{fig:HindiLLM-Tokenizer_output}. There are four sentences (three Hindi and one English) and its corresponding tokens as tokenized by our tokenizer. We can see that the words are split into multiple sub-words. For example, the word "{\dn aApkA}" is tokenized into "{\dn aApk}" and "{\dn A}"   but the word "{\dn aAp}" is a token on its own. The HindiLLM tokenizer is able to tokenize both Hindi and English sentences.

To check if the HindiLLM tokenizer makes sense or not, we have taken a 100 words paragraph written in Hindi, encoded it using our own and the default GPT-2 tokenizer. We see that our tokenizer takes 345 tokens and the default GPT-2's tokenizer takes 785 tokens to represent the same paragraph with 100 words. It takes less than half the number of tokens for our HindiLLM tokenizer. This indicates that the HindiLLM tokenizer has lower fertility score for the Hindi language. Hence, we can pass larger Hindi sentences or paragraphs to the model. This will also improve the efficiency in processing Hindi text. Therefore, creation of a new tokenizer is beneficial here.

\subsection{Unsupervised Pre-Training}
\label{training_approach-pre_training}
As mentioned in the GPT-1 \cite{radford2018improving} work, the pre-training finds a good initialization point for the model. In pre-training, the model learns about the language such as morphology and syntax. The Causal Language Modeling (CLM) is used in the unsupervised pre-training step of autoregressive models. This step gives us a base model that can be supervised fine-tuned for several downstream tasks with a relatively smaller dataset. We have trained two models, HindiLLM-Small and HindiLLM-Medium corresponding to GPT2-small and GPT2-medium respectively. We use Hugging Face's Transformers library \cite{wolf2020transformers} for pre-training. The training process involves creating a model with the configuration of corresponding GPT-2 models and training it after initializing with random weights. 

\subsubsection{HindiLLM-Small Model:}

\begin{table}
\scriptsize
    \centering
    \caption{Training Details of HindiLLM-Small Model}
    \label{table:HindiLLM-Small-training-specs}
    \begin{tabular}{|p{7cm}|p{4cm}|} \hline  
         \textbf{Specification}& \textbf{Value}\\ \hline 
         Corpus Size&19.6 GB \\ \hline  
         Number of Examples & 6,904,242 \\ \hline
         Total Train Batch Size (w. Parallel, Distributed and Accumulation) & 16 \\ \hline
         Total Optimization Steps & 431,516 \\ \hline
         Number of Trainable Parameters & 124,439,808 \\ \hline 
         Context Window Size&1024 tokens\\ \hline
         Number of Epochs & 1.45 \\ \hline 
         Optimizer&AdamW\\ \hline 
         Learning Rate&5e-05\\ \hline
         GPUs Used& 2 x NVIDIA A100-PCIE-40GB\\ \hline
         Time Required&6 days\\\hline
    \end{tabular}
\end{table}

As shown in Table \ref{table:HindiLLM-Small-training-specs}, the HindiLLM-Small model is equivalent to the GPT2-small model which has 124,439,808 trainable parameters. For training, we have taken 19.6 GB of data from our pre-training dataset. We have taken approximately half data from all corpora mentioned in Table \ref{tab:pre-training_data_description}. There are 6,904,242 examples for training a context window of 1024 tokens. We have used a batch size of 16 for training and updated the weights 431,516 times, which is equivalent to 1.45 epochs. The optimizer is a torch \cite{Paszke_PyTorch_An_Imperative_2019} based AdamW \cite{loshchilov2017decoupled} optimizer with a learning rate 5e-05. Full precision training on two NVIDIA A100-PCIE-40GB GPUs have taken 6 days including the tokenization and evaluation steps. Instead of mixed-precision \cite{micikevicius2017mixed} training, we have opted for full precision training because we have trained from scratch. 

\subsubsection{HindiLLM-Medium Model:}

\begin{table}
\scriptsize
    \centering
    \caption{Training Details of HindiLLM-Medium Model}
    \label{table:HindiLLM-Medium-training-specs}
    \begin{tabular}{|p{7cm}|p{4cm}|} \hline  
         \textbf{Specification}& \textbf{Value}\\ \hline 
         Corpus Size&37.34 GB \\ \hline  
         Number of Examples & 11,351,587 \\ \hline
         Total Train Batch Size (w. Parallel, Distributed and Accumulation) & 32 \\ \hline
         Total Optimization Steps & 354,737 \\ \hline
         Number of Trainable Parameters & 354,823,168 \\ \hline 
         Context Window Size&1024 tokens\\ \hline
         Number of Epochs & 1.24 \\ \hline 
         Optimizer&AdamW\\ \hline 
         Learning Rate&5e-05\\ \hline
         GPUs Used& 2 x NVIDIA A100-PCIE-40GB\\ \hline
         Time Required&25 days\\\hline
    \end{tabular}
\end{table}

As depicted in Table \ref{table:HindiLLM-Medium-training-specs}, the HindiLLM-Medium model is based on the same configuration as of GPT2-medium model which has 354,823,168 trainable parameters. For training, we have 37.34 GB of data as mentioned in Table \ref{tab:pre-training_data_description}. There are 11,351,587 examples for training with the same context window as HindiLLM-Small. The total optimization steps are 354,737 with a batch size of 32 which is equal to 1.24 epochs. The optimization steps are less as compared to HindiLLM-Small because we have doubled the batch size. The optimizer and learning rate are the same as HindiLLM-Small. Similar to the HindiLLM-Small model, we have used two NVIDIA A100-PCIE-40GB GPUs for performing full precision training, which took 25 days including the tokenization and evaluation steps.

\subsubsection{Performance Evaluation of Pre-trained Models:}
\begin{table}
    \centering
    \caption{Performance of the Pre-trained HindiLLM Models}
    \label{table:HindiLLM-training-performance}
    \begin{tabular}{|l|r|r|} \hline  
         \textbf{Metrics}& \textbf{HindiLLM-Small}&\textbf{HindiLLM-Medium} \\  \hline  
         Evaluation Accuracy &  0.6855  &0.7300\\  \hline
         Evaluation Loss &  1.3045  &1.0992  \\ \hline
         Perplexity &  3.6860 &3.0017 \\ \hline
         Train Loss &  1.3849  &1.2096  \\ \hline
    \end{tabular}
\end{table}

As shown in Table \ref{table:HindiLLM-training-performance}, the HindiLLM-Medium is better than HindiLLM-Small in terms of all the metrics. The larger model has higher accuracy, lower loss and lower perplexity as compared to the smaller one. The perplexity of 3.686 and 3.0017 for HindiLLM-Small and HindiLLM-Medium, respectively, assures the quality of the training. 

\subsection{Supervised Fine-Tuning}
\label{training_approach-fine_tuning}
The next step in the semi-supervised training approach is the Supervised Fine-Tuning (SFT) on discriminative or generative tasks. The fine-tuning step aligns the model with the downstream task. The previous pre-training step makes it easier to fine-tune because it has already gained knowledge about the language. Hence, we can achieve higher performance on downstream tasks even with a smaller dataset. The model can be fine-tuned for variety of tasks. Here, we have fine-tuned for seven tasks on the datasets mentioned in Section \ref{dataset-fine_tuning}. Since the model will be used on real-world downstream applications, the SFT and evaluation of the resultant model are the crucial steps. 

We have used deepspeed \cite{rasley2020deepspeed} library for fine-tuning efficiently using multi-GPU setup. The SFT is done in bfloat16 precision format with a learning rate of 5e-6.

\section{Performance Evaluation}
\label{evaluation}
In this section, the performances on the downstream tasks are evaluated. We have done SFT for a variety of downstream tasks to get a detailed performance measure of the HindiLLM models. The results are compared with other models to validate its improvement.

\subsection{Public Classification Dataset}
\label{evaluation-public_classification}

\begin{table}
\centering
\caption{Classification Accuracy on Public Dataset}
\label{tab:supervised_fine-tuning_result_1}
\begin{tabular}{| l | r | r | r |}
\hline
\textbf{Model} & \textbf{IITP-Movie} & \textbf{IITP-Product} & \textbf{BBC-Article} \\
\hline
HindiLLM-Small & 70.51 & 76.63 & 71.29 \\
\hline
HindiLLM-Medium & \textbf{78.34} & \textbf{79.31} & \textbf{81.04} \\
\hline
FT-W & 41.61 & 58.32 & 72.29 \\
\hline
FT-WC & 44.52 & 57.17 & 67.44 \\
\hline
IndicFT & 45.81 & 61.57 & 77.02 \\
\hline
IndicBERT-Base & 59.03 & 71.32 & 74.60 \\
\hline
mBERT & 56.77 & 74.57 & 60.55 \\
\hline
XLM-R & 61.61 & 78.97 & 75.52 \\
\hline
INLP & 45.81 & 63.48 & 74.25 \\
\hline
iNLTK & 57.74 & 75.71 & 78.75 \\
\hline
GPT-3.5 Turbo Zero-shot & 68.17 & 68.20 & 56.41\\
\hline
GPT-3.5 Turbo Few-shot & 66.45 & 72.92 & 49.63\\
\hline

\end{tabular}

\end{table}

As shown in Table \ref{tab:supervised_fine-tuning_result_1}, we compare various models on public classification datasets. We have accuracy scores from Wikipedia (FT-W) \cite{bojanowski2017enriching}, Wiki+CommonCrawl (FT-WC) \cite{qi2020stanza}, IndicFT \cite{kakwani2020indicnlpsuite}, IndicBERT \cite{kakwani2020indicnlpsuite}, mBERT \cite{pires2019multilingual}, XLM-R \cite{ruder2019unsupervised}, INLP \cite{kunchukuttan2020ai4bharat} and iNLTK \cite{arora2020inltk} models. Also, we have obtained scores from the GPT-3.5 Turbo model. For the GPT-3.5 Turbo model, we have considered zero-shot prompting and few-shot prompting. We have performed prompt engineering to find the best system prompt and in the case of few-shot prompting, we have given five random examples from the training data. We can observe that the HindiLLM-Medium model surpasses all the models on all three datasets: IITP-Movie, IITP-Product, and BBC-Article public classification datasets. The HindiLLM-Small model comes second in the case of IITP-Movie dataset. It does not perform well on the BBC-Article dataset. For IITP-Movie dataset, we see an improvement of $2.34\%$ and $10.17\%$ in HindiLLM-Small and HindiLLM-Medium models, respectively. For IITP-Product dataset, we see a decrease of $2.34\%$ and an increase of $0.34\%$ in HindiLLM-Small and HindiLLM-Medium models, respectively. We observe an improvement of $2.29\%$ for HindiLLM-Medium model in BBC-Article dataset,  but a drop of $7.46\%$ for HindiLLM-Small. Both the zero-shot and few-shot results from GPT-3.5 Turbo are poorer than both of our models. In most cases, the result of few-shot is worse than the zero-shot. This is possibly because the given examples are confusing the model or because with a increase in the prompt length, the model is not able to analyze accurately.

\textbf{\subsection{IndicGLUE Benchmark Dataset}
\label{evaluation-indicglue}}
\begin{table}
\centering
\caption{Accuracy Score on IndicGLUE Benchmark Dataset}
\label{tab:supervised_fine-tuning_result_2}
\begin{tabular}{| l | r | r | r |}
\hline
 \textbf{Model}& \textbf{CSQA} & \textbf{WSTP} & \textbf{DM} \\
\hline
HindiLLM-Small & 38.53 & 69.85 & 78.68 \\
\hline
HindiLLM-Medium & 44.71& 77.19 & \textbf{80.48} \\
\hline
XLM-R & 30.62 & 76.92 & 79.94 \\
\hline
mBERT & 39.00 & \textbf{80.12} & 71.20 \\
\hline
IndicBERT-Base & 41.55 & 74.02 & 78.44 \\
\hline
IndicBERT-Large & 37.01 & 77.80 & NA \\
\hline
GPT-3.5 Turbo Zero-shot & 44.56 & 76.75 & 50.91 \\
\hline
GPT-3.5 Turbo Few-shot & \textbf{50.84} & 74.25 & 48.89 \\
\hline

\end{tabular}

\end{table}

Table \ref{tab:supervised_fine-tuning_result_2} shows the accuracy score achieved on the IndicGLUE benchmark dataset. 
We have considered XLM-R, mBERT and IndicBERT models for comparison along with zero-shot and few-shot prompting of GPT-3.5 Turbo. In the CSQA task, the GPT-3.5 Turbo Few-shot has the highest accuracy and our HindiLLM-Medium model has the second best accuracy which is a drop of $6.13\%$. HindiLLM-Small has the sixth best result with a drop of $12.31\%$ on this task. For the WSTP task, mBERT shows the best result whereas we see a drop of $2.93\%$ and $10.27\%$ for HindiLLM-Medium and HindiLLM-Small, respectively. 
For the DM task, HindiLLM-Medium gives the best accuracy with an improvement of $0.54\%$, followed by XLM-R and HindiLLM-Small. We do not have any score from IndicBERT-Large model on this task.

\textbf{\subsection{Comparison with GPT-2 Models}
\label{evaluation-internally}}

\begin{table}
\centering
\caption{Sentiment Analysis Comparison with Internally Fine-tuned GPT-2 Models}
\label{tab:sentiment_analysis_comparison}
\begin{tabular}{| l | r | r | r |}
\hline
\textbf{Model} & \textbf{Precison } & \textbf{Recall} & \textbf{F1-score} \\
\hline
HindiLLM-Small & 77.5 & 75.22 & 76.34 \\
\hline
HindiLLM-Medium & \textbf{80.6} & \textbf{79.18} & \textbf{79.88} \\
\hline
GPT2-Small & 56.77 & 44.47 & 49.87 \\
\hline
GPT2-Medium & 62.84 & 63.89 & 63.36 \\
\hline

\end{tabular}

\end{table}

\begin{table}
\centering
\caption{Natural Language Inference Comparison with Internally Fine-tuned GPT-2 Models}
\label{tab:nli_comparison}
\begin{tabular}{| l | r | r | r |}
\hline
\textbf{Model} & \textbf{Precison } & \textbf{Recall} & \textbf{F1-score} \\
\hline
HindiLLM-Small & 97.16 & 97.24 & 97.20 \\
\hline
HindiLLM-Medium & \textbf{98.03} & \textbf{99.18} & \textbf{98.60} \\
\hline
GPT2-Small & 70.23 & 69.61 & 69.92 \\
\hline
GPT2-Medium & 70.75 & 69.35 & 70.04 \\
\hline

\end{tabular}

\end{table}

We have fine-tuned the default GPT-2 \cite{radford2019language} along with our models for the Sentiment Analysis dataset (a combination of IITP-Movie and IITP-Product datasets) and BBC-NLI dataset. We have used the same train-test split for fine-tuning both GPT-2 and HindiLLM model. The fine-tuning process was the same for both kinds of models. From the results shown in Table \ref{tab:sentiment_analysis_comparison} and Table \ref{tab:nli_comparison}, it is evident that there is a huge improvement in the performance on Hindi downstream tasks using HindiLLM models. Even the smaller HindiLLM-Small model surpasses the scores of GPT2-Medium models by a large margin.

\textbf{\subsection{Machine Translation Dataset}}
\label{evaluation-translation}
\begin{table}
\centering
\caption{Human Evaluation Criteria}
\label{tab:translation_evaluation_criteria}
\begin{tabular}{| l | r |}
\hline
\textbf{Quality} & \textbf{Rating} \\
\hline
Excellent & 4 \\
\hline
Good & 3 \\
\hline
Understandable & 2 \\
\hline
Barely understandable & 1 \\
\hline
Incomprehensible & 0 \\
\hline
\end{tabular}
\end{table}

In machine translation, we have considered both English-to-Hindi and Hindi-to-English translation tasks using the same set of train and test data. For machine translation, we have only considered the HindiLLM-Medium model. The smaller model will struggle a bit here because it is a generative task. Since the Hindi language is morphologically rich, it is unfair to use traditional metrics such as BLEU and METEOR. Hence, we have performed human evaluation of the translations using the criteria mentioned in Table \ref{tab:translation_evaluation_criteria}.

\begin{table}
\centering
\caption{HindiLLM-Medium model on Machine Translation Dataset}
\label{tab:translation_result}
\begin{tabular}{| l | r | r |}
\hline
\textbf{Metric} & \textbf{Hindi to English} & \textbf{English to Hindi} \\
\hline
Score 4 & 6.89\% & 5.91\% \\
\hline
Score 3 & 28.83\% & 44.93\% \\
\hline
Score 2 & 28.67\% & 29.47\% \\
\hline
Score 1 & 31.40\% & 17.81\% \\
\hline
Score 0 & 4.21\% & 1.88\% \\
\hline
Mean Score & 2.03 & 2.35 \\
\hline
\end{tabular}
\end{table}

Table \ref{tab:translation_result} shows the performance of the machine translation task. We have shown the probability distribution of the scores. The English to Hindi task performs slightly better than the Hindi to English task. We have a small portion of data for score 0, which is good. But for score 4 as well we have a small portion of data. Even when the score is 3, the translation quality is good. We see that a substantial portion of data for the Hindi to English task and a major portion of data for the English to Hindi task have a score of 3. It indicates that the translation quality is good. The mean scores are above average. Considering that we have used limited English data in pre-training, the results are promising.

\section{Conclusion}
\label{conclusion}

In this paper, we train a tokenizer and two auto-regressive models of different sizes for the Hindi language written in Devanagari script. We check the validity of the models by comparing the results on multiple downstream tasks. By looking at the performances, we conclude that the pre-trained models are well-trained to handle a variety of downstream tasks.

As we see the performance evaluation of downstream tasks, in most of the cases our HindiLLM-Medium model shows the best results. The HindiLLM-Small model lags because of its smaller size and smaller pre-training data. It is clear from the results that HindiLLM will contribute to solving real-world problems, especially the HindiLLM-Medium model. We also note that our HindiLLM models perform better than the fine-tuned GPT-2 and prompt engineering GPT-3.5 Turbo model. This implies that training a language-specific model will result in better performance for that particular language even with a smaller model. Also, it is evident from the results that training a larger model on a larger dataset will result in a better performing model.

Even though the models show impressive results, there is scope for further improvements. The number of epochs during training can be further increased. The training can be performed with enhanced data like the text from the books. Training a larger model on larger pre-training data will always result in a better model. Our model has limited knowledge of English since a small portion of our dataset comprises English data. The English data can be increased to enhance the bilingual capability. Furthermore, adding a few more supervised fine-tuning tasks can add to the assurance of the quality of the model. 


In the future, we plan to create models by combining Hindi, Romanized Hindi (Hinglish), and English data. Adding Hinglish data can make it more relevant in day-to-day applications. We see frequent use of Hinglish these days instead of Hindi and it is rapidly gaining popularity. A few such examples are comment sections, posts and messages on social networking sites. Further, adding more English texts to the pre-training data will make it bilingual. This will increase its usability in tasks like machine translation and tasks that contain a mix of Hindi and English data.

%
%
%

%


\begin{thebibliography}{10}
\providecommand{\url}[1]{\texttt{#1}}
\providecommand{\urlprefix}{URL }
\providecommand{\doi}[1]{https://doi.org/#1}

\bibitem{abadji2022towards}
Abadji, J., Suarez, P.O., Romary, L., Sagot, B.: Towards a cleaner document-oriented multilingual crawled corpus. arXiv preprint arXiv:2201.06642  (2022)

\bibitem{achiam2023gpt}
Achiam, J., Adler, S., Agarwal, S., Ahmad, L., Akkaya, I., Aleman, F.L., Almeida, D., Altenschmidt, J., Altman, S., Anadkat, S., et~al.: Gpt-4 technical report. arXiv preprint arXiv:2303.08774  (2023)

\bibitem{anil2023palm}
Anil, R., Dai, A.M., Firat, O., Johnson, M., Lepikhin, D., Passos, A., Shakeri, S., Taropa, E., Bailey, P., Chen, Z., et~al.: Palm 2 technical report. arXiv preprint arXiv:2305.10403  (2023)

\bibitem{arora2020inltk}
Arora, G.: inltk: Natural language toolkit for indic languages. arXiv preprint arXiv:2009.12534  (2020)

\bibitem{bojanowski2017enriching}
Bojanowski, P., Grave, E., Joulin, A., Mikolov, T.: Enriching word vectors with subword information. Transactions of the association for computational linguistics  \textbf{5},  135--146 (2017)

\bibitem{conneau2019unsupervised}
Conneau, A., Khandelwal, K., Goyal, N., Chaudhary, V., Wenzek, G., Guzm{\'a}n, F., Grave, E., Ott, M., Zettlemoyer, L., Stoyanov, V.: Unsupervised cross-lingual representation learning at scale. arXiv preprint arXiv:1911.02116  (2019)

\bibitem{dai2019transformer}
Dai, Z., Yang, Z., Yang, Y., Carbonell, J., Le, Q.V., Salakhutdinov, R.: Transformer-xl: Attentive language models beyond a fixed-length context. arXiv preprint arXiv:1901.02860  (2019)

\bibitem{el2019ccaligned}
El-Kishky, A., Chaudhary, V., Guzm{\'a}n, F., Koehn, P.: Ccaligned: A massive collection of cross-lingual web-document pairs. arXiv preprint arXiv:1911.06154  (2019)

\bibitem{howard2018universal}
Howard, J., Ruder, S.: Universal language model fine-tuning for text classification. arXiv preprint arXiv:1801.06146  (2018)

\bibitem{hu2021lora}
Hu, E.J., Shen, Y., Wallis, P., Allen-Zhu, Z., Li, Y., Wang, S., Wang, L., Chen, W.: Lora: Low-rank adaptation of large language models. arXiv preprint arXiv:2106.09685  (2021)

\bibitem{jiang2023mistral}
Jiang, A.Q., Sablayrolles, A., Mensch, A., Bamford, C., Chaplot, D.S., Casas, D.d.l., Bressand, F., Lengyel, G., Lample, G., Saulnier, L., et~al.: Mistral 7b. arXiv preprint arXiv:2310.06825  (2023)

\bibitem{kakwani2020indicnlpsuite}
Kakwani, D., Kunchukuttan, A., Golla, S., Gokul, N., Bhattacharyya, A., Khapra, M.M., Kumar, P.: Indicnlpsuite: Monolingual corpora, evaluation benchmarks and pre-trained multilingual language models for indian languages. In: Findings of the Association for Computational Linguistics: EMNLP 2020. pp. 4948--4961 (2020)

\bibitem{kunchukuttan2020ai4bharat}
Kunchukuttan, A., Kakwani, D., Golla, S., Bhattacharyya, A., Khapra, M.M., Kumar, P., et~al.: Ai4bharat-indicnlp corpus: Monolingual corpora and word embeddings for indic languages. arXiv preprint arXiv:2005.00085  (2020)

\bibitem{kunchukuttan2017iit}
Kunchukuttan, A., Mehta, P., Bhattacharyya, P.: The iit bombay english-hindi parallel corpus. arXiv preprint arXiv:1710.02855  (2017)

\bibitem{lan2019albert}
Lan, Z., Chen, M., Goodman, S., Gimpel, K., Sharma, P., Soricut, R.: Albert: A lite bert for self-supervised learning of language representations. arXiv preprint arXiv:1909.11942  (2019)

\bibitem{loshchilov2017decoupled}
Loshchilov, I., Hutter, F.: Decoupled weight decay regularization. arXiv preprint arXiv:1711.05101  (2017)

\bibitem{micikevicius2017mixed}
Micikevicius, P., Narang, S., Alben, J., Diamos, G., Elsen, E., Garcia, D., Ginsburg, B., Houston, M., Kuchaiev, O., Venkatesh, G., et~al.: Mixed precision training. arXiv preprint arXiv:1710.03740  (2017)

\bibitem{niyogi2024paramanu}
Niyogi, M., Bhattacharya, A.: Paramanu: A family of novel efficient indic generative foundation language models. arXiv preprint arXiv:2401.18034  (2024)

\bibitem{owen2024komodo}
Owen, L., Tripathi, V., Kumar, A., Ahmed, B.: Komodo: A linguistic expedition into indonesia's regional languages. arXiv preprint arXiv:2403.09362  (2024)

\bibitem{Paszke_PyTorch_An_Imperative_2019}
Paszke, A., Gross, S., Massa, F., Lerer, A., Bradbury, J., Chanan, G., Killeen, T., Lin, Z., Gimelshein, N., Antiga, L., Desmaison, A., Kopf, A., Yang, E., DeVito, Z., Raison, M., Tejani, A., Chilamkurthy, S., Steiner, B., Fang, L., Bai, J., Chintala, S.: {PyTorch: An Imperative Style, High-Performance Deep Learning Library}. In: Wallach, H., Larochelle, H., Beygelzimer, A., d'Alché Buc, F., Fox, E., Garnett, R. (eds.) Advances in Neural Information Processing Systems 32. pp. 8024--8035. Curran Associates, Inc. (2019), \url{http://papers.neurips.cc/paper/9015-pytorch-an-imperative-style-high-performance-deep-learning-library.pdf}

\bibitem{pires2019multilingual}
Pires, T., Schlinger, E., Garrette, D.: How multilingual is multilingual bert? arXiv preprint arXiv:1906.01502  (2019)

\bibitem{qi2020stanza}
Qi, P., Zhang, Y., Zhang, Y., Bolton, J., Manning, C.D.: Stanza: A python natural language processing toolkit for many human languages. arXiv preprint arXiv:2003.07082  (2020)

\bibitem{radford2018improving}
Radford, A., Narasimhan, K., Salimans, T., Sutskever, I., et~al.: Improving language understanding by generative pre-training  (2018)

\bibitem{radford2019language}
Radford, A., Wu, J., Child, R., Luan, D., Amodei, D., Sutskever, I., et~al.: Language models are unsupervised multitask learners. OpenAI blog  \textbf{1}(8), ~9 (2019)

\bibitem{rasley2020deepspeed}
Rasley, J., Rajbhandari, S., Ruwase, O., He, Y.: Deepspeed: System optimizations enable training deep learning models with over 100 billion parameters. In: Proceedings of the 26th ACM SIGKDD International Conference on Knowledge Discovery \& Data Mining. pp. 3505--3506 (2020)

\bibitem{ruder2019unsupervised}
Ruder, S., S{\o}gaard, A., Vuli{\'c}, I.: Unsupervised cross-lingual representation learning. In: Proceedings of the 57th Annual Meeting of the Association for Computational Linguistics: Tutorial Abstracts. pp. 31--38 (2019)

\bibitem{touvron2023llama}
Touvron, H., Martin, L., Stone, K., Albert, P., Almahairi, A., Babaei, Y., Bashlykov, N., Batra, S., Bhargava, P., Bhosale, S., et~al.: Llama 2: Open foundation and fine-tuned chat models. arXiv preprint arXiv:2307.09288  (2023)

\bibitem{uppal2020two}
Uppal, S., Gupta, V., Swaminathan, A., Zhang, H., Mahata, D., Gosangi, R., Shah, R., Stent, A.: Two-step classification using recasted data for low resource settings. In: Proceedings of the 1st Conference of the Asia-Pacific Chapter of the Association for Computational Linguistics and the 10th International Joint Conference on Natural Language Processing. pp. 706--719 (2020)

\bibitem{vaswani2017attention}
Vaswani, A., Shazeer, N., Parmar, N., Uszkoreit, J., Jones, L., Gomez, A.N., Kaiser, {\L}., Polosukhin, I.: Attention is all you need. Advances in neural information processing systems  \textbf{30} (2017)

\bibitem{de2020good}
de~Vries, W., Nissim, M.: As good as new. how to successfully recycle english gpt-2 to make models for other languages. arXiv preprint arXiv:2012.05628  (2020)

\bibitem{wenzek2019ccnet}
Wenzek, G., Lachaux, M.A., Conneau, A., Chaudhary, V., Guzm{\'a}n, F., Joulin, A., Grave, E.: Ccnet: Extracting high quality monolingual datasets from web crawl data. arXiv preprint arXiv:1911.00359  (2019)

\bibitem{wolf2020transformers}
Wolf, T., Debut, L., Sanh, V., Chaumond, J., Delangue, C., Moi, A., Cistac, P., Rault, T., Louf, R., Funtowicz, M., et~al.: Transformers: State-of-the-art natural language processing. In: Proceedings of the 2020 conference on empirical methods in natural language processing: system demonstrations. pp. 38--45 (2020)

\end{thebibliography}
\end{document}